# Adaptive Neuro Fuzzy Networks based on Quantum Subtractive Clustering

Ali Mousavi*, Mehrdad Jalali and Mahdi Yaghoubi

*Abstract*—Data mining techniques can be used to discover useful patterns by exploring and analyzing data andit's feasible to synergistically combine machine learning tools to discover fuzzy classification rules.In this paper, an adaptive neuro fuzzy network with TSK fuzzy type and an improved quantum subtractive clustering has been developed. Quantum clustering (QC) is an intuition from quantum mechanics which uses Schrödinger potential and time-consuming gradient descent method. The principle advantage and shortcoming of QC is analyzedandbased on its shortcomings, an improved algorithm through a subtractive clustering method is proposed. Cluster centers represent a general model with essential characteristics of data which can be use as premise part of fuzzy rules.The experimental results revealed that proposed Anfis based on quantum subtractive clustering yielded good approximation and generalization capabilities and impressive decrease in the number of fuzzy rules and network output accuracy in comparison with traditional methods.

## I. INTRODUCTION

The last decade has seen an explosive growth in the generation and collection of data, advances in data mining and automation of predictor systems. Hence a lot of new techniques and tools that can intelligently and automatically assist in transforming this data into useful knowledge have been proposed.

An Adaptive Neuro Fuzzy Inference System (ANFIS) is a framework based on the concepts of fuzzy systems which have been improved by artificial neural networks. So Anfis tries to integrate advantages of fuzzy systems and artificial neural networks. This means that the neural networks present learning capabilities to fuzzy systems via a connectionist structure. Moreover fuzzy systems provide a framework to work with human knowledge and the uncertain world [1, 2].

Methods of data clustering are usually based on geometric or probabilistic considerations. The problem of unsupervised learning of clusters based on locations of points in data-space is in general ill-defined [3, 6]. Hence intuition based on other fields of study may be useful in formulating new heuristic procedures. There are a lot of clustering methods which leads to generate adequate numbers of fuzzy rules. The cluster centers represent a general model with essential characteristics of data. Each cluster center can be use as premise part of a fuzzy rule [2, 6].

Here we use a model based on tools that are borrowed from quantum mechanics. The quantum clustering introduced by Horn [3] is a new and unique clustering method that is an extension of idea inherent to scale-space and support-vector clustering. In addition, this is represented by the Schrödinger equation, which is a potential function that can analytically be derived from a probability function. The effectiveness of this clustering has been demonstrated on pattern recognition [1, 3].

Nonetheless the Quantum Clustering (QC) method has some drawbacks. It is difficult to determine the number of valid cluster centers, because this QC severely depends on a one-variable parameter representing the scale of its Gaussian kernel, and it is hard to deal with a high dimension of input space so QC method is modified in a simple manner [1, 3].

Also a subtractive clustering method has been used to determine the adequate number of cluster centers. This is a density based clustering that proposed by C h i u  i n  1 9 9 4  [ 7 ] .

Therefore we propose a new method to construct an adaptive neuro fuzzy network with a TSK fuzyy type. Also we use a modified QC to determine the premise part of fuzzy rules. Moreover subtractive clustering method is applied to determine the optimal number of clusters. We performed a learning method by a hybrid learning scheme using back propagation (BP) and a least-square estimator (LSE). The experiments used the well-known automobile mile-per-gallon (MPG) prediction dataset which consists of 392 records. In this example, six input variables are composed of a car's cylinder number, displacement, horsepower, weight, acceleration, and model year. The output variable to be predicted by the six input variables is the fuel consumption of the automobile.

This paper is organized as follows: In section II, we describe basics of Anfis. The Quantum Clustering method, our modified Quantum clustering and subtractive clustering are presentedin section III and followed Section IV gives theexperimental results. Finally, we conclude the paper in sectionV.

*Ali Mousavi is with Department of Artificial Intelligence, Faculty of Engineering, Mashhad Branch, Islamic Azad University,Mashhad, Iran.
(E-mail: mousavi@mshdiau.ac.ir).

Mehrdad Jalali is with department of Artificial Intelligence, Assistant Professor (PhD),Mashhad Branch, Islamic Azad University, Mashhad, Iran.
(E-mail: mehrdadjalali@ieee.org)

Mahdi Yaghoubi is with department of Electrical Engineering, Assistant Professor (PhD),Mashhad Branch, IslamicAzad University, Mashhad, Iran.
(E-mail: yaghoubi@mshdiau.ac.ir)

## II. BASICS OF ANFIS

ANFIS is a neuro-fuzzy system developed byJang[2]. It has a feed-forward neural networkstructure where each layer is a neuro-fuzzy system .In this section, we describe the basic architecture, and learning rules of the Anfis. The Anfis structure identification involves two phases: 1) structure identification and 2) parameter identification. The former is related to determining the number of fuzzy if–then rules and a proper partition of the input space. The latter is concerned with the learning of model parameters, such as membership functions andlinear coefficients. As shown in Fig. 1, this network is composed of five layers that introduced by Jang et al [2]. Nevertheless the Anfis suffers from curse of dimensionality that the number of extracted fuzzy rules increases exponentially due to grid partitioning of input data. In this paper we use scatter one instead to find adequate number of rules through new proposed clustering method. For a zero and first order Sugeno Fuzzy model common rules are as following respectively. Consider the Anfis model has n inputs[1, 4].

$R^i$: If $p_1 is A_1^i$ And $p_2$ is $A_2^i$ And... And $p_n$ is $A_n^i$,
 then $y^i = a_0^i$ (1)

$R^i$: If $p_1 is A_1^i$ And $p_2$ is $A_2^i$ And... And $p_n$ is $A_n^i$,
 then $y^i = a_0^i + a_1^i p_1 + a_2^i p_2 + ... + a_n^i p_n$ (2)

Here $A_j^i$ the linguistic value from the *j*th input variable $p_j$ in the *i*th fuzzy rule. Also $a_j^i$ is a constant. Now we describe duty of each layer in the following briefly:

Layer 1, the first layer of this network involves $n$ nodes which has a membership function as weight of links. The output determines the firing strength of premise part of associated proposition of a rule. A Gaussian membership function with two parameters$(\mu, \sigma)$is used where these parameters obtain from clustering phase.

Layer 2, each node in Layer 2 provides the firing strength of the rule by means of multiplication operator. It performs AND operation.

$$w_i = \prod_{j=1}^{n} A_j^i(p_j), \quad i = 1,2,...,n \quad (3)$$

Every node in this layer computes the multiplication of the input values and gives the product as the output according to the above equation.

Layer 3 is the normalization layer which normalizesthe strength of all rules according to the following equation.

$$\overline{w_i} = \frac{w_i}{\sum_{i=1}^{r} w_i} \quad (4)$$

Where $w_i$ is the firing strength of the *i*th rule which iscomputed in Layer 2. Node $i$ computes the ratio ofthe *i*thrule's firing strength to the sum of all rules' firing strength.

Layer 4, everynode in the fourth layer computes a product operation (And) between thenormalized firing strength from layer 3 and consequentpart of corresponding rule. It means output of this layer is $\overline{w_i}.y^i$ that shows firing strength of each rule.

Layer 5, the single node of this layer aggregates all of incoming inputs.Here $\hat{y}$ is the final predicted output of the network.

$$\hat{y} = \sum_{i=1}^{r} \overline{w_i} y^i \quad (5)$$

Because of the overall output is obtained through a summation operator, we don't need to complicated defuzzification process.

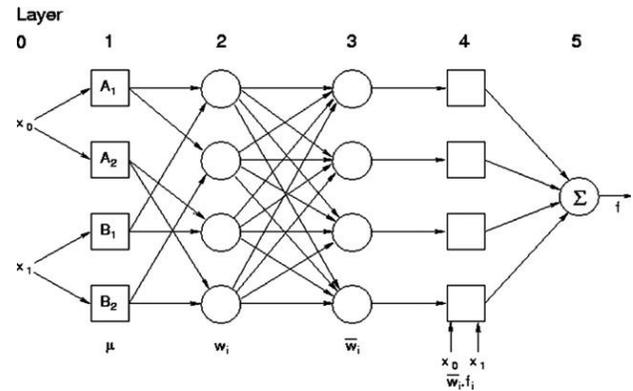

Fig. 1.Anfis structure.

## III. RULE EXTRACTION VIA CLUSTERING

### A. Quantum Clustering

Clustering is the procedure that classifies the set of physicalabstract objects to several clusters composed of similar objects.Methods of data clustering are usually based on geometric or probabilistic considerations [6]. We can consider intuition based on other fields of study to formulating new clustering procedures. Quantum clustering introduced by Horn [3] is a novel method from quantum mechanics that research an operator in Hilbert space expressed by Schrödinger equation, the solution is wave-function. When the wave-function is given, they can work out the potential function by Schrödinger equation, which will determine the distribution of particle, and the cluster centers are the k minimum in the potential. According to the quantum theory, particle with lowerpotential vibrate less, and relatively stable, so we can treat this kind of particle as the cluster centers, and then distribute data points to the associated clusters [5].

In QC algorithm, wave function in equation (6)is used to describe the sample's distribution and produce the estimator parameter of Parzen Window.

$$\Psi(x) = \sum_{i=1}^{n} e^{-(x-x_i)^2/2\sigma^2} \quad (6)$$

Where $x_i$ is the $i$th data point.

In the QC algorithm, we use the k minima of Schrödinger potential to determine the location of the cluster centers. This potential is part of the Schrödinger equation (7), for which $\psi(x)$ is a solution.

$$H\psi \equiv \left(-\frac{\sigma^2}{2}\nabla^2 + V(x)\right)\psi(x) = E\psi(x) \quad (7)$$

Where $H$ is the Hamiltonian, $V$ is the potential energy, $E$ is the energy eigenvalue and $E = d/2$. When $V(x)$ is positive definite, we obtain $V(x) \geq 0$. Hence, $E$ is defined as follows:

$$E = -min\frac{\frac{\sigma^2}{2}\nabla^2\psi}{\psi} \quad (8)$$

The eigenvalue $E$ of Schrödinger's equation is the lowest eigenfunctions of the operator $H$ representing the ground state. Moreover, when the minima of $V(x)$ are defined as the cluster centers, the assignments of data points to clusters are obtained by a gradient descent algorithm allowing auxiliary point variables $z_k(0) = x_k$, $k = 1, 2, \ldots, N$, to follow dynamics as follows:

$$y_i(t + \nabla t) = y_i(t) - \eta(t)\nabla V(y_i(t)) \quad (9)$$

Here $z_k$ is used as the pre-clusters, and $\eta(t)$ is the learning rate [3].

### B. Modified QC

The proposed QC algorithm is a classic quantum clustering algorithm, but it does have some defects. It's difficult to find the learning turns in the gradient descent iterative procedure which seeks the potential minimum. Accordingly that effects on the number of cluster centers and cluster accuracy. Furthermore the gradient descent is a time consuming iterative procedure. The proposed method improves QC to solve above problems by elimination of gradient descent procedure and using a subtractive clustering algorithm. The proposed method is illustrated as following:

First, calculate the vector of potential $V$ for all data samples according to (8) equation. Second, use subtractive clustering explained in section C to obtain optimal number of cluster centers as parameter, $k$. Third, the cluster centers are calculated through following stages:

1- Sort samples in vector $V$ by ascending.
2- Choose first $N$th associated data points as cluster centers.

### C. Subtractive Clustering

Subtractive method belongs to density based method was proposed by Chiu in 1994. Chiu suggested an improved version of the mountain method, which is proposed by Yager and Filev [7]. Here we use SC to obtain optimal number of cluster centers.

The SC algorithm assumes each data point is a potential cluster center. Consider a collection of $N$ data point in $m$-dimensional space. Then calculate a density measure for data points as follows [1]:

$$D_k = \sum_{k'=1}^{N} \exp\left(-\frac{\|z_k - z_{k'}\|^2}{r_\alpha/2)^2}\right) \quad (10)$$

Where $r_\alpha$ is a constant and $z_k(k = 1,2,3,\ldots,N)$ is $k$th data point. A data point with many neighboring data points will have a high density value. Thus, data points with a high value of potential are more suited to be the potential cluster centers. A data point with highest density value is selected as the first cluster center. Then the density measure for all of data points will be update by following equation:

$$D_k = D_k - D_{v1}\exp\left(-\frac{\|z_k - z_{v1}\|^2}{r_b/2)^2}\right) \quad (11)$$

Where $r_b = 1.5r_\alpha$ and $v_1$ is center of selected cluster. The data points around the first cluster center will have reduced density measures. Thus, these data points never selected as the next cluster center. The process continues until a sufficient number of cluster centers were obtained [5].

## IV. EXPERIMENTAL RESULT

Generally effective partitioning of input space can reduce number of fuzzy rules and increase learning speed of Anfis. In this paper we use a quantum subtractive clustering to determine fuzzy rules, and then a modified Anfis is applied to data set to predict the output. The experiments used the well-known automobile mile-per-gallon (MPG) prediction dataset which consists of 392 records. In this example, six input variables are composed of car's cylinder number, displacement, horsepower, weight, acceleration, and model year. The output variable to be predicted by the six input variables is the fuel consumption of the automobile. At first records with missing values has been eliminated, then we normalize data to [0,1] interval. Data set divided into two partition; train data and test data, according to even record and odd records from original dataset, respectively. Here we use training data set to construct and learn the model while test data is used to validate the model.

At first number of cluster centers ($k$) is obtained by subtractive clustering, then we use modified QC to find $k$ cluster centers. Finally, we construct the Anfis model and efficient number of fuzzy rules will be generated. Here we have obtained 20 rules. The network output predicts the fuel consumption of automobile.

We have compared our new Quantum Subtractive Clustering method with traditional QC applied in Anfis [1].

Fig. 2(a) shows comparison results between the desired and traditional QC model outputs for test data. Here horizontal axis shows number of test data and vertical axis determines desired and model fuel consumption. Also Fig .2(b) shows comparison diagrams between the desired and our model outputs for test data set.

As shown in Fig. 2, it is obvious that the difference between desired and proposed model output is less than previous method. As the result, proposed method has better approximation and generalization capability.

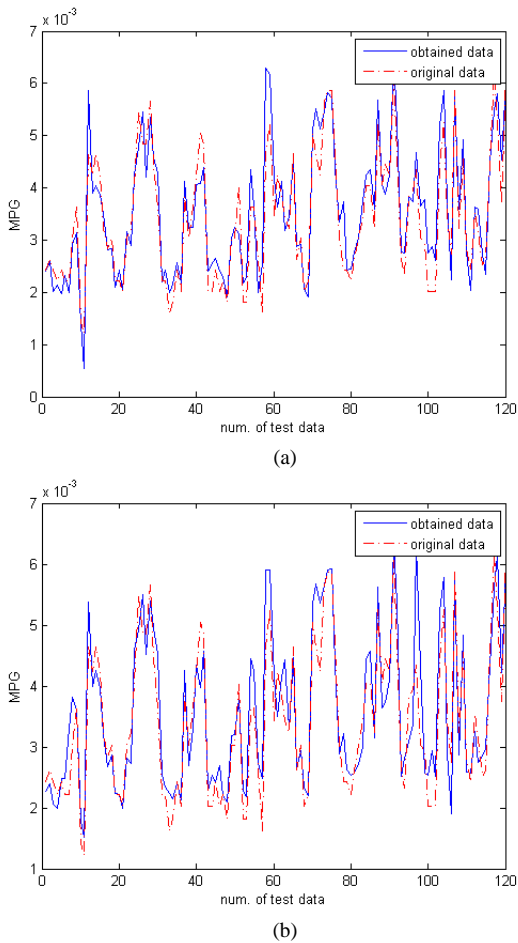

Fig. 2.Approximation and generalizationcapability in previous method (a) and proposed method (b).

Fig. 3 shows the root mean square error (RMSE) for both new and previous methods during 10 epochs respectively. As Fig. 3(a) is shown, at the beginning there is a lot of difference between test and train error rate, because of initial epochs of model, but it decreases gradually and reaches to a reasonable value in comparison with error rate in previous method. The calculated average testing error for new method is 0.0029 while it's 0.0056 for previous method. It shows a considerable improvement in comparison with before.

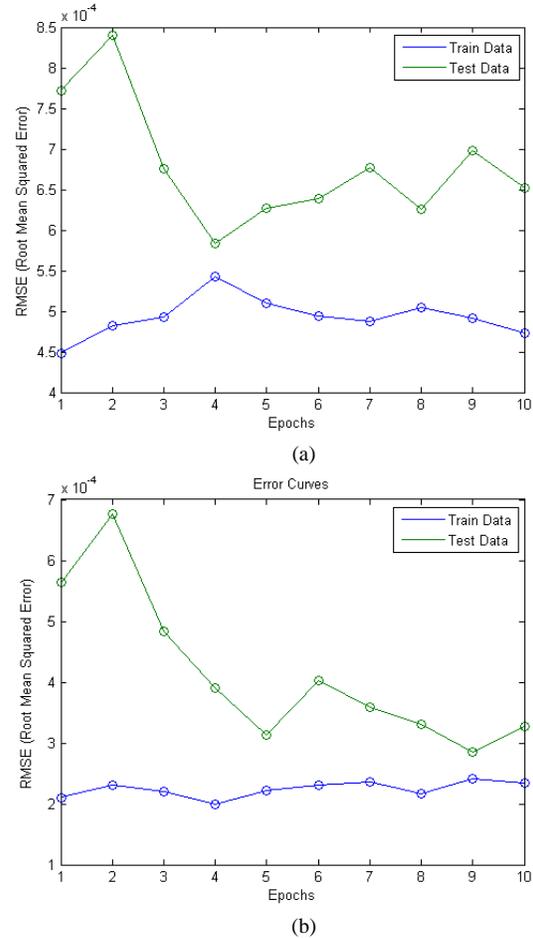

Fig. 3.Comparison of RMSE in the training and test data in proposed method (a) and previous method (b).

## V. CONCLUSION

An Adaptive Neuro Fuzzy Network with a TSK fuzzy type combined with an improved quantum subtractive clustering method to obtain appropriate number of fuzzy rules is proposed. The subtractive clustering, a density based algorithm, is used to determine number of cluster centers. Moreover a modified quantum clustering, an idea from quantum mechanics is applied to obtain cluster centers. Cluster centers represent a general model with essential characteristics of data which can be use as premise part of fuzzy rules. It caused impressive decrease in number of fuzzy rules and network accuracy. Finally we construct our model to predict fuel consumption in MPG dataset.

The experimental results showed the proposed method has good approximation and generalization capabilities compared with traditional one. Moreover our new quantum based subtractive clustering is more accurate and faster. As the result, fuzzy rules obtained from quantum based clustering improve our model undoubtedly.

Our future work is replacing hard quantum subtractive clustering method with fuzzy quantum subtractive clustering approach to increase the accuracy of method and deal with real problems in our world where fuzzy concept is a solution.